\documentclass[letterpaper]{article} % DO NOT CHANGE THIS
\usepackage{aaai2026}  % DO NOT CHANGE THIS
\usepackage{times}  % DO NOT CHANGE THIS
\usepackage{helvet}  % DO NOT CHANGE THIS
\usepackage{courier}  % DO NOT CHANGE THIS
\usepackage[hyphens]{url}  % DO NOT CHANGE THIS
\usepackage{graphicx} % DO NOT CHANGE THIS
\usepackage{natbib}  % DO NOT CHANGE THIS AND DO NOT ADD ANY OPTIONS TO IT
\usepackage{caption} % DO NOT CHANGE THIS AND DO NOT ADD ANY OPTIONS TO IT
\usepackage{algorithm}
\usepackage{algorithmic}
\usepackage{newfloat}
\usepackage{listings}
\DeclareCaptionStyle{ruled}{labelfont=normalfont,labelsep=colon,strut=off}
\floatstyle{ruled}
\newfloat{listing}{tb}{lst}{}
\floatname{listing}{Listing}
\usepackage{tabularx}
\usepackage{lipsum}
\usepackage{xcolor}
\usepackage{latexsym}
\usepackage{amsmath}
\usepackage{amsfonts}
\usepackage{dsfont}
\usepackage{multirow}
\usepackage{booktabs}
\usepackage{colortbl}
\usepackage{subcaption}
\usepackage{times}
\usepackage{helvet}
\usepackage{courier}
\usepackage{arydshln}
\usepackage{graphicx}
\usepackage{array}
\usepackage{colortbl}
\usepackage[most]{tcolorbox}

\definecolor{cardframe}{HTML}{2C3E50}
\definecolor{humanback}{HTML}{EAF2FB}
\definecolor{humanrule}{HTML}{3B7DD8}
\definecolor{botback}{HTML}{EFF7EE}
\definecolor{botrule}{HTML}{4A9D5B}
\definecolor{sysback}{HTML}{F2F2F2}
\definecolor{sysrule}{HTML}{888888}
\definecolor{sectionblue}{RGB}{220,230,241}

\newtcolorbox{cardframebox}[1]{enhanced jigsaw, breakable,
colback=white, colframe=cardframe, boxrule=0.6pt, arc=3pt,
left=6pt, right=6pt, top=8pt, bottom=2pt,
fonttitle=\bfseries\footnotesize\sffamily, coltitle=white, title={#1},
attach boxed title to top left={xshift=8pt, yshift=-2pt},
boxed title style={colback=cardframe, arc=2pt, boxrule=0pt},
toprule at break=0pt, bottomrule at break=0pt}

\usepackage{arydshln}

\title{Benchmarking LLMs' Mathematical Reasoning \\ with Unseen Random Variables Questions}

\author{
    Zijin~Hong\textsuperscript{\rm 1}\thanks{Both authors contributed equally to this research.},
    Hao~Wu\textsuperscript{\rm 2}\footnotemark[1],
    Su~Dong\textsuperscript{\rm 1},
    Junnan~Dong\textsuperscript{\rm 3},
    Yilin~Xiao\textsuperscript{\rm 1},
    Yujing~Zhang\textsuperscript{\rm 1},
    Zhu~Wang\textsuperscript{\rm 1}
    Feiran~Huang\textsuperscript{\rm 4},
    Linyi~Li\textsuperscript{\rm 5},
    Hongxia~Yang\textsuperscript{\rm 1},
    Xiao~Huang\textsuperscript{\rm 1}\thanks{Corresponding author.}
}
\affiliations {
    \textsuperscript{\rm 1}The Hong Kong Polytechnic University,\,Hong Kong SAR,\, China \\
    \textsuperscript{\rm 2}University of Electronic Science and Technology of China,\,Chengdu,\,China \\
    \textsuperscript{\rm 3}Tencent Youtu Lab,\,Shanghai,\,China \\
    \textsuperscript{\rm 4}Beihang University,\,Beijing,\,China \\
    \textsuperscript{\rm 5}Simon Fraser University,\,Burnaby,\,Canada \\
    
    zijin.hong@connect.polyu.hk, wh.pyjnqd@gmail.com, xiao.huang@polyu.edu.hk
}

\begin{document}

\maketitle

\begin{abstract}
Recent studies have raised significant concerns regarding the reliability of current mathematical benchmarks, highlighting key limitations such as simplistic design and potential data contamination that undermine evaluation accuracy.
Consequently, developing a reliable benchmark that effectively evaluates large language models' (LLMs) genuine capabilities in mathematical reasoning remains a critical challenge.
To address these concerns, we propose \textbf{\textsc{RV-Bench}}, a novel evaluation methodology for \textbf{\textsc{Bench}}marking LLMs with \textbf{\textsc{R}}andom \textbf{\textsc{V}}ariables in mathematical reasoning.
Specifically, we develop question-generating functions to produce random variable questions (RVQs), whose background content mirrors the original benchmark problems, but with randomized variable combinations, rendering them ``unseen'' to LLMs.
Models must completely understand the inherent question pattern to correctly answer RVQs with diverse variable combinations. 
Thus, an LLMs' genuine reasoning capability is reflected through its accuracy and robustness on \textsc{RV-Bench}. 
We conducted extensive experiments on over 30 representative LLMs across more than 1,000 RVQs. 
Our findings reveal that LLMs exhibit a proficiency imbalance between encountered and ``unseen'' data distributions. 
Furthermore, \textsc{RV-Bench} reveals that proficiency generalization across similar mathematical reasoning tasks is limited, but we verified that it can still be effectively elicited through test-time scaling.
\end{abstract}

\begin{links}
    \link{Code}{https://github.com/DEEP-PolyU/RV-Bench}
    \link{Extended version}{https://arxiv.org/abs/2501.11790}
\end{links}

\section{Introduction}
The emergence of LLMs has led to impressive results across a wide range of applications, including machine translation~\cite{zhang2023prompting}, text summarization~\cite{liu2024learning}, and question answering~\cite{kamalloo2023evaluating}. 
With advancements in LLMs' reasoning capabilities~\cite{huang2023towards}, their performance on complex real-world challenges such as code generation~\cite{chen2021evaluating,hong2024next}, planning~\cite{huang2024understanding}, and especially mathematical reasoning and computation~\cite{romera2024mathematical} has become a central focus within the LLM research community~\cite{zhao2023survey}. 
Advanced domain-specific studies on LLMs’ mathematical reasoning~\cite{ shao2024deepseekmath} further highlight their strong potential for addressing real-world challenges. 
Consequently, numerous promising methods~\cite{luo2023wizardmath} and benchmarks~\cite{zhou2025is} have been developed to further enhance and comprehensively evaluate LLMs' performance on mathematical tasks~\cite{mirzadeh2024gsm}.

\begin{figure}[!t]
    \centering
    \includegraphics[width=0.474\textwidth]{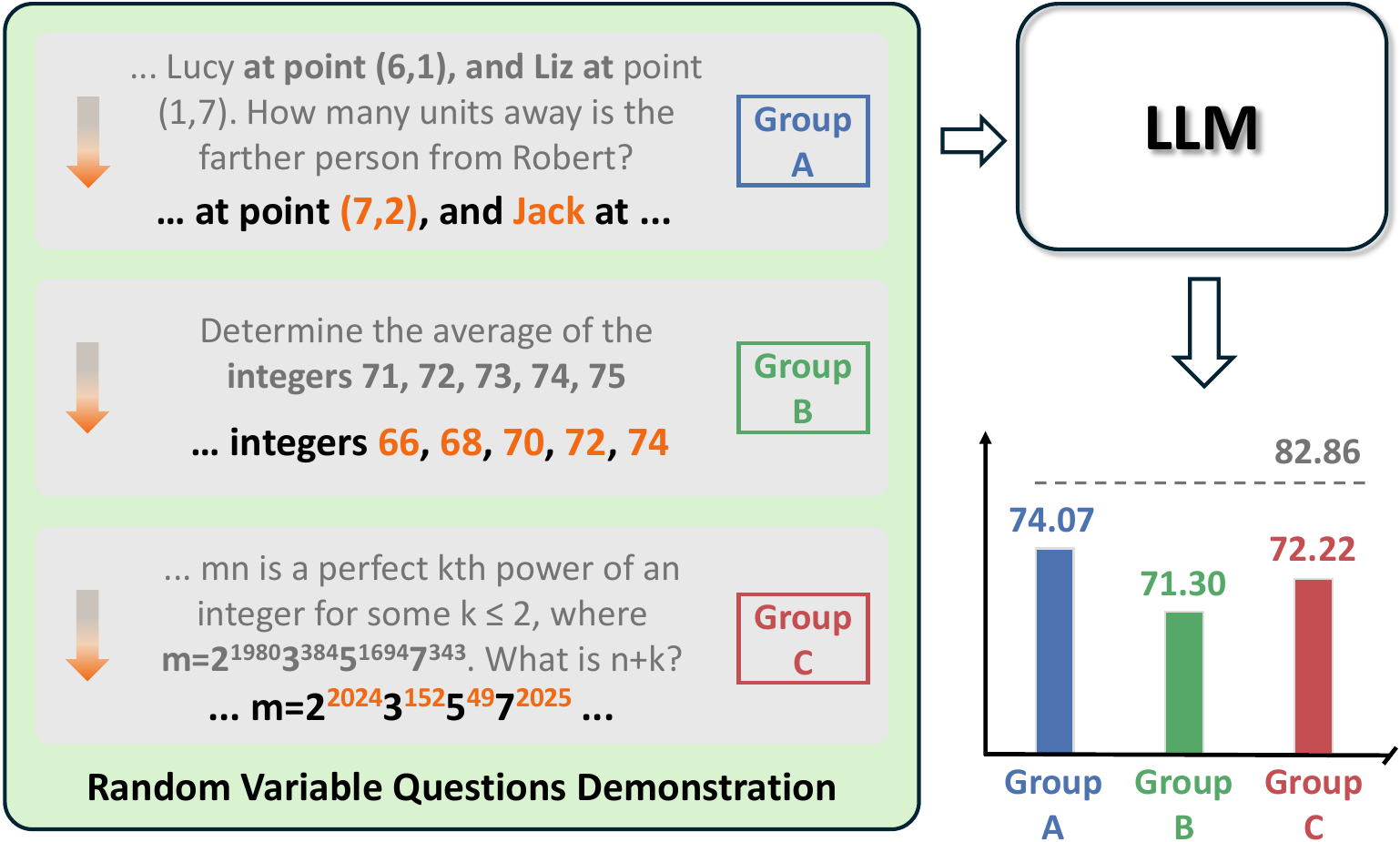} 
    \caption{When mathematical problems are presented with ``unseen'' random variables, LLMs suffer a significant accuracy drop. This discrepancy highlights the limitations of existing evaluations of the mathematical reasoning of LLMs.}
    \label{fig:introduction}
\end{figure}

However, \textbf{are existing benchmarks for LLMs' mathematical reasoning truly reliable?} 
Figure~\ref{fig:introduction} illustrates a discrepancy in the well-known MATH~\cite{hendrycks2021mathdataset} dataset. 
In our pilot experiments, powerful LLMs like GPT-4o~\cite{achiam2023gpt4} perform well on MATH problems but suffer a significant accuracy drop when answering questions with identical content but different variable combinations~\cite{mirzadeh2024gsm}, as detailed in the experimental section. 
This discrepancy raises two critical concerns about current evaluation frameworks: 
\textbf{1) Existing benchmarks may be overly simplistic for contemporary LLMs}, as they typically evaluate performance on fixed-variable problems with one-step reasoning.
LLMs may not truly understand the problems but rather ``guess'' the correct answers~\cite{dong2024clr,mirzadeh2024gsm} to achieve high performance; 
\textbf{2) Problems in widely-used benchmarks might be encountered by LLMs during training through data contamination}, enabling models to achieve high accuracy on original benchmark problems~\cite{ni2024training} without genuinely understanding the inherent question patterns.
These concerns pose a significant challenge in evaluating the genuine mathematical reasoning capabilities of LLMs~\cite{deng2024investigating}.

Advanced studies present in-depth analyses of LLMs' probabilistic modeling during the reasoning process, obscuring the fact that these models are not genuinely capable of formal reasoning~\cite{freda2023large,jiang2024peek}. 
Additionally, potential issues such as data contamination and overfitting during LLM training have been widely studied~\cite{balloccu2024leak, xu2024benchmark,mirzadeh2024gsm}, suggesting that LLMs can ``reason'' simply by memorizing and replicating the same steps.
Given that mathematics is a foundational domain applicable across a wide range of semantic scenarios, the growing popularity of publicly available datasets like GSM8K~\cite{cobbe2021gsm8k} and MATH~\cite{hendrycks2021mathdataset} increases the risk of data contamination.
Although recent studies on contamination detection~\cite{chern2024behonest, ni2024training} can signal unreliable results, they fail to reflect LLMs’ genuine performance, as data contamination occurs during pretraining and remains non-intervenable~\cite{kapoor2023leakage}.

The abovementioned phenomenon raises a critical issue: existing benchmarks may fail to accurately reflect LLMs' genuine performance on mathematical tasks~\cite{balloccu2024leak, mirzadeh2024gsm}.
In this context, \textbf{effectively evaluating LLMs' genuine mathematical reasoning capabilities remains a significant challenge}.
As a solution, this paper introduces \textbf{\textsc{RV-Bench}}, a new evaluation framework for benchmarking LLMs’ mathematical reasoning through \textbf{random variable questions (RVQs)}: diverse, ``unseen’’ questions generated with randomized variable combinations and algorithmic problems reformulated into mathematical expressions that remain out-of-distribution for LLMs' training.
\textsc{RV-Bench} provides an effective evaluation methodology that directly addresses the two concerns discussed above.

Specifically, we construct question-generating functions based on original problems from two selected mathematical data sources: the MATH~\cite{hendrycks2021mathdataset} test set and the LeetCode-Math branch. 
These functions dynamically generate instantiated questions with random variable combinations and corresponding answers.
The resulting RVQs are then collected to evaluate LLMs. 
Unlike existing mathematical benchmarks~\cite{cobbe2021gsm8k,hendrycks2021mathdataset}, \textsc{RV-Bench} comprises questions with a wide range of variable combinations, rather than fixed, static problems.
Furthermore, \textbf{\textsc{RV-Bench} provides ``unseen'' and out-of-distribution questions, enabling LLMs to demonstrate genuine mathematical reasoning capabilities} even if they have previously encountered the original problems~\cite{mirzadeh2024gsm}. 
To achieve high accuracy on \textsc{RV-Bench}, an LLM must completely understand the inherent question pattern to correctly answer RVQs, thereby effectively evaluating its genuine mathematical reasoning capabilities.
Overall, our contributions are listed as follows:

\begin{itemize}
\item We propose \textsc{RV-Bench}, a leaderboard for comprehensively evaluating LLMs' genuine mathematical reasoning capabilities using four well-designed metrics.
Our macroscopic analysis quantifies LLMs’ understanding of inherent mathematical question patterns.

\item We reveal a significant accuracy drop when LLMs solve RVQs compared to the original problems, exposing the unreliability of existing benchmarks that overlook factors such as data contamination and randomness.

\item By analyzing LLMs' accuracy and robustness on \textsc{RV-Bench}, we suggest LLMs acquire partial mathematical reasoning proficiency, which has limited generalization but can be effectively elicited through test-time scaling.
\end{itemize}

\section{Related Work}
The rapid advancement of LLMs has triggered the development of benchmarks for evaluating their general and domain-specific capabilities~\cite{chang2024survey}. 
General-purpose benchmarks like MMLU~\cite{hendrycks2021mmlu}, GLUE~\cite{wang2018glue} assess broad tasks such as question answering and natural language understanding, while domain-specific datasets such as GSM8K~\cite{cobbe2021gsm8k} and MATH~\cite{hendrycks2021mathdataset} target mathematical reasoning. 
With the emergence of math-specialized models like Qwen-Math~\cite{yang2024qwen25math} and DeepSeek-Math~\cite{shao2024deepseekmath}, performance on existing math benchmarks has reached near-perfect. 
However, recent advanced studies like GSM-Symbolic~\cite{mirzadeh2024gsm} and PAL~\cite{gao2023pal} reveal that such improvements may arise from pattern replication rather than truly understanding and reasoning. 
Our proposed \textsc{RV-Bench} addresses this concern by introducing mathematical questions with random variables, challenging LLMs to beyond memorization and better evaluate their genuine reasoning capabilities. Further related works are in the extended version. 

\section{\textsc{RV-Bench} Construction}
\begin{figure*}[!t]
    \centering
    \includegraphics[width=\textwidth]{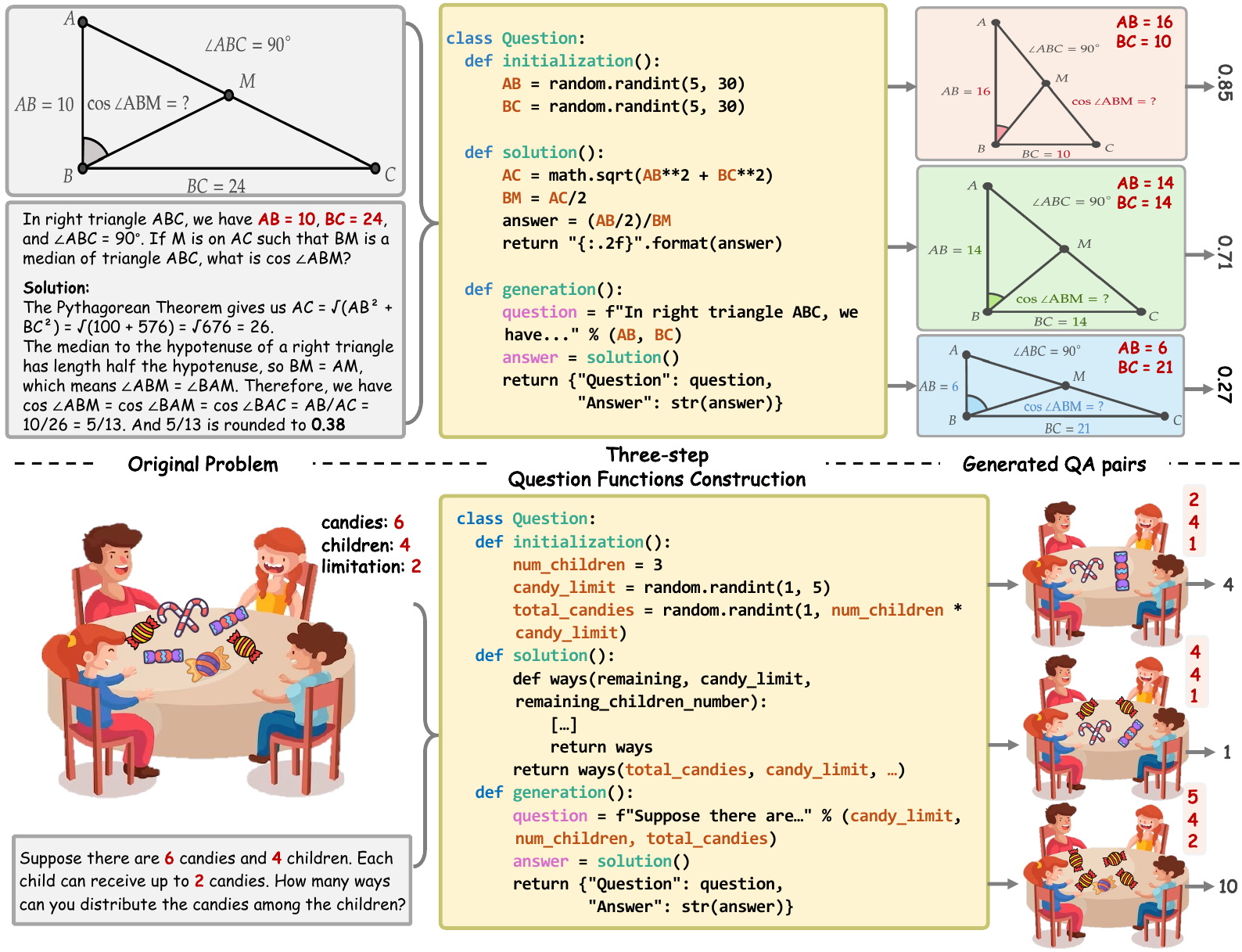} 
    \caption{The workflows of \textsc{RV-Bench} are illustrated for the MATH (top) and LeetCode (bottom) data sources. Each question-generating function (\texttt{Question}) comprises three modules: First, the \texttt{initialization} (\texttt{init}) module randomizes a variable combination. Next, the \texttt{solution} (\texttt{sol}) module computes the corresponding answer. Finally, the \texttt{generation} (\texttt{gen}) module outputs the instantiated question and its answer, forming a QA pair for the questions utilized in \textsc{RV-Bench} evaluation.}
    \label{fig:rv-bench}
\end{figure*}

Figure~\ref{fig:rv-bench} provides a sketch workflow for \textsc{RV-Bench} construction from both the MATH and LeetCode data sources. 
In this section, we will provide a brief introduction to the data sources and question-generating function form of \textsc{RV-Bench}, and detail the process in the extended version.

\subsection{Data Sources \& Sampling Strategy}
\label{sec:data_sources}
Two selective data sources are used for constructing \textsc{RV-Bench}: the MATH test set and the LeetCode-Math branch. 
MATH~\cite{hendrycks2021mathdataset} is a well-known dataset that covers 12,500 challenging mathematics problems targeted at high-school mathematics competitions. 
Following the pre-processing settings of the MATH by PRM800K~\cite{lightman2024lets}, we construct 120 question-generating functions based on randomly and uniformly selected problems from the test split. 
LeetCode is a well-recognized platform providing algorithmic problems for users to practice coding skills~\cite{tristan2024leetcode}. 
LeetCode-Math is a branch that includes algorithmic problems whose content is designed based on mathematical reasoning and computation. 
Our motivation for selecting LeetCode as our data source comes from its original focus on coding problems. 
By transforming these problems into mathematical expressions, we posit that the resulting formulations are unlikely to have been encountered during LLMs’ training, rendering them out-of-distribution and effectively ``unseen'' to the models.
Through a careful review of each problem, we construct 130 question-generating functions by manually reformatting the content with random variables, selected uniformly at random. 
Consequently, the question-generating functions in \textsc{RV-Bench} are constructed based on the respective data sources, maintaining similar distributions of difficulty, type, and bias as the original dataset.

\subsection{Question Functions \& Difficulty Control}
As illustrated in Figure~\ref{fig:rv-bench}, a question function consists of three modules: \texttt{init}, \texttt{sol}, and \texttt{gen}. 
These modules are responsible for instantiating the random variables, solving the RVQs, and generating the RVQ-A pairs for \textsc{RV-Bench}, respectively.
To enable a fair comparison between RVQs and original problems, we implement a strict difficulty calibration for each question function. 
Specifically, both the numerical complexity and conceptual difficulty of each RVQ are manually controlled to ensure consistency, since empirical studies~\cite{gao2023pal} indicated that increasing numerical magnitude can lead to accuracy drop. 
However, it primarily reflects the LLMs' limited arithmetic capability, rather than their genuine mathematical reasoning in RVQs.
This tailored calibration minimizes confounding effects arising from difficulty and inconsistency in RVQs. 
Further details and analysis are also provided in the extended version.

\section{Experimental Setups}
\paragraph{Datasets.}
\label{sec:datasets}
Overall, \textsc{RV-Bench} comprises 230 question functions, evenly split between MATH and LeetCode-Math (115 each).
To compare LLMs under random variables and original settings, we sample the corresponding original problems for each question function, denoted as \textbf{\textsc{MATH-Sp}} and \textbf{\textsc{LeetCode-Sp}}, respectively.
For each question function, we construct an RVQ group with \textbf{five} RVQs instantiated using varied variable combinations.
In total, 1,150 RVQs are generated from 230 RVQ groups based on the manually designed question functions in \textsc{RV-Bench}.
These RVQs are evenly split by data source into \textbf{\textsc{MATH-RV}} and \textbf{\textsc{LeetCode-RV}}, both of which are used throughout the paper.
More details are listed in the extended version.

\paragraph{Evaluation Metrics.}
We define four metrics for the evaluation process. 
Given a set of RVQ groups $\mathcal{Q}_{\textsc{RV}} = \left\{ \mathcal{G}^{(i)} \right\}_{i=1}^{m}$, let $\mathcal{Q}_{\textsc{Sp}}$ denote the corresponding set of original problem groups.
Each RVQ group generated from the $i$-th question function is denoted as $\mathcal{G}^{(i)} = \left(q^{(i)}_{j}\right)^{n}_{j=1} \in \mathcal{Q}_{\textsc{RV}}$, where $n$ is the number of RVQs in the group $\mathcal{G}^{(i)}$.
The original problem corresponding to $\mathcal{G}^{(i)}$ is denoted as $q^{(i)}_{\textsc{Sp}} \in \mathcal{Q}_{\textsc{Sp}}$.
Let $\hat{a}^{(i)}_j$, $a^{(i)}_j$, $\hat{a}^{(i)}_{\textsc{Sp}}$, and $a^{(i)}_{\textsc{Sp}}$ denote the predicted and ground-truth answers for $q^{(i)}_j$ and $q^{(i)}_{\textsc{Sp}}$, respectively. 
We further define $N_{\mathcal{G}^{(i)}}$ as the number of correctly answered RVQs associated with $\mathcal{G}^{(i)}$:

\noindent \textbf{1) Exact Match Accuracy (Acc)}: Measures the correctness of the answer for each RVQ through strict string matching:
\begin{equation}
    \text{Acc} = \frac{\sum^{m}_{i}\sum^{|\mathcal{G}^{(i)}|}_{j}\mathds{1}\left(\hat{a}^{(i)}_j = a^{(i)}_j\right)}{m\cdot n}.
\end{equation}

\noindent \textbf{2) Group Accuracy@$n$ ($\text{GA@$n$}$)}: Indicates that all $n$ generated RVQs are answered correctly in $\mathcal{G}^{(i)}$:
\begin{equation}
\text{GA@$n$} = \frac{\sum_{i=1}^{m} \mathds{1}\left(\forall q^{(i)}_j \in \mathcal{G}^{(i)}, \hat{a}^{(i)}_j = a^{(i)}_j\right)}{m}.
\end{equation}

\noindent \textbf{3) Complete Ratio (CR)}: Evaluate whether the original problem is answered correctly and at least 80\% of the RVQs are also correct, as a measure of the model's understanding:
\begin{equation}
\text{CR} = \frac{\sum_{i=1}^{m} \mathds{1}\left(\hat{a}^{(i)}_{\textsc{Sp}} = a^{(i)}_{\textsc{Sp}} \land N_{\mathcal{G}^{(i)}} \geq \lceil 80\% \cdot n \rceil\right)}{m}.
\end{equation}

\noindent \textbf{4) Original Only Ratio (OOR)}: Evaluates whether the original problem is answered correctly while at least 80\% of the RVQs are incorrect, indicating the proportion of cases where the model solves the original but fails on the RVQs:
\begin{equation}
\text{OOR} = \frac{\sum_{i=1}^{m} \mathds{1}\left(\hat{a}^{(i)}_{\textsc{Sp}} = a^{(i)}_{\textsc{Sp}} \land N_{\mathcal{G}^{(i)}} \le \lceil 20\% \cdot n \rceil\right)}{m}.
\end{equation}

\begin{table*}[t]\footnotesize
\centering
\setlength{\tabcolsep}{1.79pt}
\begin{tabular}{cllccccccccccc} 
\toprule
\multirow{2}{*}{\textbf{\#}} & \multirow{2}{*}{\textbf{Models}} & \multirow{2}{*}{\textbf{Size}} 
& \textbf{\textsc{MATH-Sp}} 
& \multicolumn{4}{c}{\textbf{\textsc{MATH-RV}}}
& \textbf{\textsc{LeetCode-Sp}}
& \multicolumn{4}{c}{\textbf{\textsc{LeetCode-RV}}}
& \multicolumn{1}{c}{\textbf{\textsc{RV-Bench}}} \\
\cmidrule(lr){4-4}
\cmidrule(lr){5-8}
\cmidrule(lr){9-9}
\cmidrule(lr){10-13}
\cmidrule(lr){14-14}
& & &\textbf{Acc} &\textbf{Acc} &\textbf{GA@5} &\textbf{CR} &\textbf{OOR}~($\downarrow$) &\textbf{Acc} &\textbf{Acc} &\textbf{GA@5} &\textbf{CR} &\textbf{OOR}~($\downarrow$) & \textbf{Acc} \\
\midrule[0.6pt]
1  & o3-mini            & $\sim$ & \underline{97.39} & \textbf{92.52} & \underline{82.61} & \underline{87.83} & 6.09 & \textbf{82.61} &  \textbf{77.57} & \textbf{61.74} & \textbf{67.83} & \underline{6.09}  & \textbf{85.05} \\
2  & DeepSeek-R1        & 671B   & \textbf{100.00} & \textbf{92.52} & \textbf{85.22} & \textbf{88.70} & 6.09 & \underline{80.00} & \underline{72.17} & \underline{52.17} & \underline{57.39} & \textbf{5.22}  & \underline{82.35} \\
3  & o1-mini            & $\sim$ & 90.43 & 84.00 & 67.83 & 80.87 & \underline{5.22}  & 76.52 & 66.09 & 41.74 & 51.30 & \underline{6.09}  & 75.05 \\
4  & Gemini-2.0-Pro     & $\sim$ & 92.17 & 84.17 & 71.30 & 78.26 & 8.70  & 72.17 & 60.17 & 34.78 & 42.61 & 8.70  & 72.17  \\
5  & DeepSeek-v3        & 671B   & 89.57 & \underline{85.04} & 72.17 & 76.52 & \underline{5.22} & 66.09 & 58.26 & 34.78 & 37.39 & 12.17 & 71.65  \\
6  & GLM-Zero-Preview   & $\sim$ & 92.17 & 83.13 & 65.22 & 77.39 & 6.09  & 66.96 & 60.00 & 35.65 & 44.35 & 9.57  & 71.57 \\
7  & QwQ-32B-Preview    & 32B    & 91.30 & 83.83 & 60.87 & 79.13 & \underline{5.22}  & 62.61 & 58.96 & 30.43 & 42.61 & 7.83  & 71.40 \\
8  & Claude-3.5-Sonnet  & $\sim$ & 88.70 & 80.35 & 63.48 & 73.04 & 6.09  & 70.43 & 61.39 & 35.65 & 42.61 & 8.70  & 70.87  \\
9  & Qwen2.5-Max        & $\sim$ & 88.70 & 81.39 & 63.48 & 74.78 & 6.96  & 72.17 & 58.43 & 33.04 & 42.61 & 12.17 & 69.91  \\
10 & Qwen2.5-72B-It     & 72B    & 87.83 & 81.04 & 62.61 & 76.52 & 6.09  & 66.09 & 58.43 & 29.57 & 40.00 & 10.43 & 69.74 \\
11 & Qwen2.5-32B-It     & 32B    & 90.43 & 80.00 & 61.74 & 73.91 & \textbf{4.35}  & 69.57 & 55.48 & 26.09 & 39.13 & 12.17 & 67.74  \\
12 & GLM-4-Plus         & $\sim$ & 86.09 & 77.91 & 53.91 & 71.30 & 6.96  & 66.96 & 55.30 & 26.96 & 38.26 & 14.78 & 66.61 \\
13 & o1-preview         & $\sim$ & 80.87 & 75.83 & 42.61 & 59.13 & 6.96  & 66.09 & 54.78 & 32.17 & 40.87 & 9.57  & 65.31 \\
14 & GPT-4o             & $\sim$ & 83.48 & 76.70 & 57.39 & 63.48 & 6.09  & 61.74 & 50.09 & 20.00 & 32.17 & 13.04 & 63.40 \\
15 & Phi-4              & 14B    & 77.39 & 72.00 & 53.04 & 61.74 & 8.70  & 60.00 & 54.78 & 26.96 & 34.78 & 9.57  & 63.39  \\
16 & Llama-3.3-70B-It   &  70B   & 83.48 & 74.43 & 52.17 & 62.61 & 9.57  & 60.00 & 45.57 & 18.26 & 22.61 & 15.65 & 60.00  \\
17 & Qwen2.5-7B-It      & 7B     & 81.74 & 71.65 & 52.17 & 60.00 & 8.70  & 53.91 & 46.78 & 20.87 & 26.09 & 13.04 & 59.22 \\
18 & Qwen2.5-Math-It    & 7B     & 87.83 & 72.70 & 51.30 & 62.61 & 12.17 & 54.78 & 37.91 & 10.43 & 17.39 & 14.78 & 55.31 \\
19 & Qwen2.5-3B-It      & 3B     & 82.61 & 67.65 & 43.48 & 60.00 & 8.70  & 43.48 & 37.04 & 12.17 & 19.13 & 14.78 & 52.35 \\
20 & Llama-3.1-70B-It   & 70B    & 73.04 & 62.78 & 39.13 & 50.43 & 9.57  & 57.39 & 40.35 & 14.78 & 23.48 & 15.65 & 51.57 \\
21 & Gemma-2-27B-It     & 27B    & 66.09 & 59.13 & 34.78 & 46.96 & 6.09  & 43.48 & 35.65 & 13.04 & 17.39 & 13.91 & 47.39  \\
22 & Phi-3-medium-4k-It & 14B    & 64.35 & 53.04 & 24.35 & 35.65 & 11.30 & 50.43 & 37.22 & 8.70  & 19.13 & 13.91 & 45.13  \\
23 & Yi-1.5-Chat        & 34B    & 64.35 & 50.96 & 21.74 & 31.30 & 11.30 & 38.26 & 33.74 & 8.70  & 13.04 & 13.04 & 42.35  \\
24 & Phi-3-mini-4k-It   & 3.8B   & 63.48 & 50.26 & 26.09 & 37.39 & 14.78 & 41.74 & 34.26 & 9.57  & 16.52 & 11.30 & 42.26  \\
25 & Qwen2.5-7B-Base    & 7B     & 66.96 & 53.22 & 25.22 & 36.52 & 13.04 & 48.70 & 31.13 & 7.83  & 12.17 & 21.74 & 42.18 \\
26 & Gemma-2-9B-It      & 9B     & 65.22 & 51.30 & 30.43 & 36.52 & 13.04 & 38.26 & 29.91 & 5.22  & 12.17 & 13.91 & 40.61 \\
27 & GPT-3.5 Turbo      & $\sim$ & 60.00 & 48.35 & 20.87 & 30.43 & 11.30 & 35.65 & 31.48 & 9.57  & 14.78 & 12.17 & 39.92 \\
28 & Mathstral-7B       & 7B     & 59.17 & 45.22 & 19.13 & 29.57 & 14.78 & 35.65 & 28.70 & 6.96  & 12.17 & 11.30 & 36.96  \\
29 & Llama-3.1-8B-It    & 8B     & 57.39 & 46.43 & 25.22 & 30.43 & 16.52 & 33.04 & 27.13 & 6.96  & 10.43 & 15.65 & 36.78  \\
30 & DeepSeek-Math-It   & 7B     & 59.13 & 48.17 & 18.26 & 33.04 & 11.30 & 29.57 & 24.70 & 6.09  & 7.83  & 12.17 & 36.44 \\
31 & Mixtral-8x7B-It-v0.1    & 46.7B  & 44.35 & 33.22 & 11.30 & 13.91 & 17.39 & 41.74 & 27.65 & 6.09  & 9.57  & 20.00 & 30.44 \\
32 & Llama-3.2-3B-It    & 3B     & 50.43 & 36.70 & 15.65 & 22.61 & 14.78 & 38.26 & 23.83 & 5.22  & 9.57  & 20.00 & 30.27 \\
33 & Llama-3.1-8B-Base & 8B     & 39.13 & 24.52 & 6.09  & 12.17 & 17.39 & 34.78 & 21.57 & 5.22  & 7.83  & 16.52 & 23.05 \\
\bottomrule
\end{tabular}

\caption{The \textsc{\textbf{RV-Bench}} leaderboard for various LLMs includes RVQs from \textbf{\textsc{MATH-RV}} and \textbf{\textsc{LeetCode-RV}}. Rankings are based on \textbf{\textsc{RV-Bench} Acc (overall accuracy)}, which measures the overall exact match accuracy across all RVQs in both \textsc{MATH-RV} and \textsc{LeetCode-RV}. For intuitive comparison, we also report the accuracy on the original problems, listed under \textsc{MATH-Sp} and \textsc{LeetCode-Sp}. The best and second-best results in each column are highlighted in bold and underlined, respectively. A tilde ($\sim$) in the ``Size'' column indicates that the model is proprietary and its size is not publicly disclosed. For the proposed evaluation metrics: \textbf{Acc}, \textbf{GA@5}, and \textbf{CR}, higher values are better; and for \textbf{ORR} ($\downarrow$), lower values are better.}
\label{tab:main_results}
\end{table*}

\paragraph{Implementations.}
Following LLaMA-3~\cite{dubey2024llama3}, we use 4-shot prompting from Minerva~\cite{lewkowycz2022solving} as few-shot examples for inference on \textsc{MATH-RV} and \textsc{MATH-Sp}. 
Similarly, for \textsc{LeetCode-RV} and \textsc{LeetCode-Sp}, we randomly select 4 problems out of \textsc{LeetCode-RV} and manually craft step-by-step solutions to serve as the few-shot examples.  
All the experiments with open-source LLMs are conducted on an NVIDIA server with 8 A100 GPUs, while proprietary LLMs are accessed via APIs provided by their respective official platforms. 

\paragraph{Model Selection.}
The selected models span a diverse range of LLMs, covering various sizes and families to enable comprehensive evaluation across multiple dimensions. 
We include open-source LLMs~\cite{dubey2024llama3}, math-specific models~\cite{yang2024qwen25math}, proprietary LLMs~\cite{achiam2023gpt4}, and large reasoning models (LRMs)~\cite{guo2025deepseekr1} in \textsc{RV-Bench} evaluation. The detailed list of models is provided in the extended version.

\section{\textsc{RV-Bench} Learderboard}
Table~\ref{tab:main_results} summarizes the performance of various LLMs on the proposed \textsc{RV-Bench}. 
As expected from the metric definitions, the performance of a given LLM typically follows the order: Acc $\geq$ CR $\geq$ GA@$5$. 
Specifically, higher Acc and GA@$5$ indicate stronger performance on RVQs and greater consistency across RVQ groups, respectively. 
A higher CR reflects the model’s more complete understanding of the reasoning process underlying both the original problem and its RV variants. 
Correspondingly, a higher OOR suggests that while the model may answer the original problem correctly, it fails to capture the underlying problem structure, leading to poor generalization on the RVQs.

LLMs are expected to demonstrate higher Acc, CR, and GA@$5$, and are preferably to have lower OOR. 
\textit{Models that meet this expectation are recognized as having complete question pattern understanding and possessing genuine mathematical reasoning capabilities.}
Furthermore, the generally lower GA@$5$ suggests that while models can solve individual instances correctly, they struggle to maintain consistency across various variable combinations. 
To mitigate the potential impact of numerical complexity on the RVQs' difficulty, we compared the distribution of computational errors in the RVQs with those in the original problems, as reported in the extended version. 
The results show that under the Random Variables setting, LLMs do not experience a significant increase in computational errors. 
\textit{Accordingly, we ignore the impact of numerical complexity in the subsequent experimental analysis and conclusions in this paper.}

\begin{figure*}[!t]
    \centering
     \includegraphics[width=1.0\textwidth]{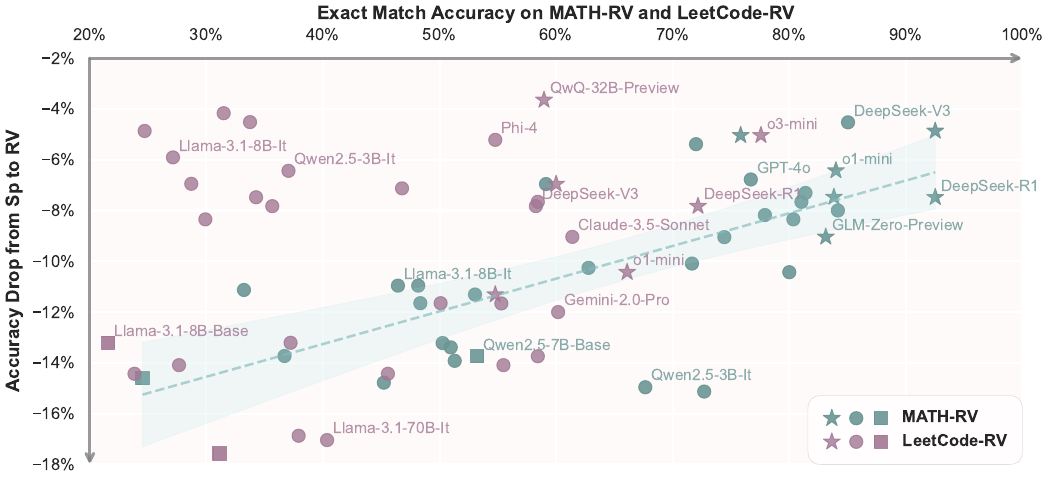}
    \caption{The accuracy drop from original problems to their corresponding RVQs is illustrated. Green data points represent the drop in accuracy from \textsc{MATH-Sp} to \textsc{MATH-RV}, while pink data points represent the drop from \textsc{LeetCode-Sp} to \textsc{LeetCode-RV}. All vertical axis values are computed as the direct difference in accuracy between the original problems and their corresponding RVQs. Different types of LLMs are indicated by distinct marker shapes, and several representative models are annotated by name for clarity. A dotted regression line is fitted using the \textsc{MATH-RV} data points, and the shaded region denotes the corresponding 95\% confidence interval. The Pearson correlation coefficient for the green data points is $r_{\textsc{M}} = -0.72$, indicating a strong negative correlation, while that for the pink data points is $r_{\textsc{L}} = -0.14$ with no clear relationship.}
    \label{fig:accuracy_drop}
\end{figure*}

In detail, o3-mini and DeepSeek-R1 lead significantly in performance on \textsc{RV-Bench}, highlighting their exceptional mathematical reasoning capabilities. 
Additionally, proprietary LRMs such as o1-mini and GLM-Zero-Preview demonstrate reliable performance. 
The open-source LRM QwQ-32B also achieves promising accuracy, surpassing that of renowned advanced LLMs such as GPT-4o and Claude-3.5. 
Large-scale chatLLMs, including Gemini-2.0-Pro, DeepSeek-V3, and Claude-3.5-Sonnet, achieve strong results, further supporting the benefits of scaling model size.
In contrast, other open-source LLMs tend to exhibit mediocre accuracy. 
In summary, the performance of LLMs on \textsc{RV-Bench} remains closely correlated with their underlying mathematical reasoning capabilities.

\subsection{Macroscopic Analysis of \textsc{RV-Bench}}
\label{sec:macroscopic_analysis}
We further advanced the analysis from a macroscopic perspective, considering the LLMs' accuracy on both RVQs from \textsc{MATH-RV} and \textsc{LeetCode-RV}, as well as the original problems from \textsc{MATH-Sp} and \textsc{LeetCode-Sp}. 
The CR and OOR metrics, reported in Table~\ref{tab:main_results}, evaluate the model’s understanding of question patterns by verifying the consistency of accuracy. Specifically, a higher accuracy correlates with a higher CR. 
Leading models, such as o3-mini and DeepSeek-R1, achieve nearly 90\% CR, demonstrating that they fully comprehend most of the inherent question patterns that can effectively handle the associated RVQs. 
For well-performing models, CR and Acc remain largely consistent, suggesting that only a small subset of question patterns eludes complete understanding.
Conversely, LLMs with lower performance and fewer parameters exhibit a higher OOR and a greater variance between CR and Acc.

Since CR and OOR indicate both complete and insufficient understanding behaviors based on accuracy inconsistencies, we further quantify the degree of the LLMs’ patterns understanding for comprehensive analysis when LLMs correctly answer the original problems. 
Specifically, we assign a pattern understanding score $S$ to each RVQ group $\mathcal{G}^{(i)}$:
\begin{equation}
    S_{\mathcal{G}^{(i)}} = 
\begin{cases} 
1, & N_{\mathcal{G}^{(i)}} \geq \lceil 0.8 \cdot n \rceil \\
0, & N_{\mathcal{G}^{(i)}} \leq \lceil 0.2 \cdot n \rceil \\
0.5, & \text{otherwise}
\end{cases}\,.
\end{equation}
Different values of $S_{\mathcal{G}^{(i)}}$ reflect varying degrees of LLMs' understanding of the question pattern corresponding to $q^{(i)}_{\textsc{Sp}}$. 
These degrees are categorized as \textbf{complete understanding} $\left(S_{\mathcal{G}^{(i)}} = 1\right)$, \textbf{partial understanding} $\left(S_{\mathcal{G}^{(i)}} = 0.5\right)$, and \textbf{collapsed understanding} $\left(S_{\mathcal{G}^{(i)}} = 0\right)$ of $q^{(i)}_{\textsc{Sp}}$, respectively.

Due to space considerations, the detailed visualization of the average frequency of various understanding behaviors and the average pattern understanding score of all the LLMs is provided in the Appendix.
We directly turn to the corresponding conclusion.
Based on that analysis, we identify that models with an average score below 0.6 demonstrate significant inconsistency in their question pattern understanding.
In other words, these models do not perform genuine mathematical reasoning capability on \textsc{RV-Bench}.

What can be concluded from the previous observation is that: the performance of nearly all LLMs on \textsc{MATH-RV} is significantly better than their performance on \textsc{LeetCode-RV}.
One possible reason for this discrepancy is the higher task difficulty and complexity of \textsc{LeetCode-RV} and \textsc{LeetCode-Sp}. 
Beyond this, we introduce another explanation based on our findings: \textbf{\textit{the mathematical reasoning capabilities of LLMs partially depend on the data distribution involved in their training, which does not generalize well across mathematical reasoning tasks.}}
Following our motivation for selecting LeetCode-related data, which is primarily utilized for enhancing coding skills and kept ``unseen'' for mathematical reasoning tasks.  
For questions in \textsc{MATH-RV}, although these RVQs remain new to the LLMs, it is highly likely that they have encountered MATH training sets with same-source problems within the same distribution to enhance their mathematical reasoning capabilities. 
Through this, LLMs can develop specific proficiency in MATH-related data. 
However, such proficiency is relatively scarce on LeetCode. 
Deducing from the performance variance, this proficiency does not generalize well, even when directly applied to similar mathematical reasoning tasks.

\subsection{Accuracy Dropping in RVQs}
\label{sec:accuracy_dropping}
Figure~\ref{fig:accuracy_drop} illustrates the accuracy drop of various LLMs when transitioning from answering the original problems in \textsc{MATH-Sp} and \textsc{LeetCode-Sp} to solving the RVQs in \textsc{MATH-RV} and \textsc{LeetCode-RV}. 
Each data point in the scatter plot represents the accuracy drop of a specific LLM on a particular data distribution. 
Significantly, all models exhibited varying degrees of accuracy drop introduced by random variable setting, ranging from 4\% to 16\%. 
The prevalence of this accuracy dropping supports our previous concern: the existing benchmark design is overly simplistic for current LLMs. 
We consider that \textbf{\textit{matching a single answer only for a fixed problem is unreliable, as it may neglect influences such as data contamination and inherent randomness, potentially introducing bias into the final results.}}
In our proposed \textsc{RV-Bench}, replacing variables in mathematical problems can lead to significant accuracy deviations.

For the observed data points corresponding to different question distributions, we intuitively fit a line using the MATH-related data points. 
When computing the correlation coefficient between the accuracy on \textsc{MATH-RV} and the accuracy drop (measured as the difference between \textsc{MATH-Sp} and \textsc{MATH-RV}), we obtain $r_{\textsc{M}} = -0.72$, indicating a strong negative correlation: models with lower performance tend to exhibit larger accuracy drops.
In other words, higher accuracy on \textsc{MATH-RV} implies better robustness and consistency across varying variable combinations in RVQs. 
By contrast, the correlation coefficient calculated for the LeetCode-related data points is $r_{\textsc{L}} = -0.14$, suggesting no clear relationship between the model’s accuracy on \textsc{LeetCode-RV} and its robustness or consistency.

Similarly, we observe that the consistency and robustness of LLMs on RVQs in \textsc{LeetCode-RV} are significantly weaker than those in \textsc{MATH-RV}. 
Beyond the possible explanation of varying task difficulty, we extend the hypothesis introduced at the end of the previous section: \textbf{\textit{the robustness and consistency of LLMs in mathematical reasoning are also partially dependent on data distribution.}}
Proficiency within a specific data distribution does not generalize well in terms of robustness and consistency to other, even similar, mathematical reasoning tasks.
In conclusion, we unify these observations into a potential underlying explanation for the inconsistency: \textbf{\textit{LLMs obtain certain proficiency in mathematical reasoning through training, but this proficiency is partially distribution-dependent. While it may apply to similar questions within the same distribution, it does not generalize reliably across others.}}
This raises an important question: does such distribution-dependent proficiency truly reflect genuine mathematical reasoning capability?

\subsection{Test-time Scaling Elicits Proficiency}
The previous section introduced two potential explanations for the inconsistency in model accuracy and robustness across different data distributions: (1) variation in task difficulty, and (2) the distribution-dependent proficiency of LLMs.
This kind of proficiency presents a critical issue: LLMs tend to replicate reasoning patterns encountered within a familiar distribution, even when faced with problems from a different, ``unseen'' distribution~\cite{mirzadeh2024gsm}, leading to biased single-turn reasoning.
To further investigate this phenomenon, we extend our experimental setting by applying test-time scaling, which enables LLMs to attempt each question multiple times~\cite{brown2024large}.
Specifically, we evaluate the LLMs' performance using $\text{pass@$k$}$ metrics~\cite{chen2021evaluating}.
For each mathematical question, the LLM generates $P$ independent answers. For $1 \leq k \leq P$, the $\text{pass@$k$}$ metric is defined as:
\begin{equation}
    \text{pass@$k$} = \operatorname{{\mathbb{E}}_{\text{Questions}}} \left[ 1 - \frac{{\binom{P-c}{k}}} {\binom{P}{k}} \right],
\end{equation}
where $c$ denotes the number of correctly answered questions. We set $P$ = 256 and re-evaluate two selected LLMs: Llama-3.2-3B-It and GPT-3.5 Turbo using the $\text{pass@$k$}$ metric.

\begin{figure}[!t]
    \centering
    \includegraphics[width=1.0\linewidth]{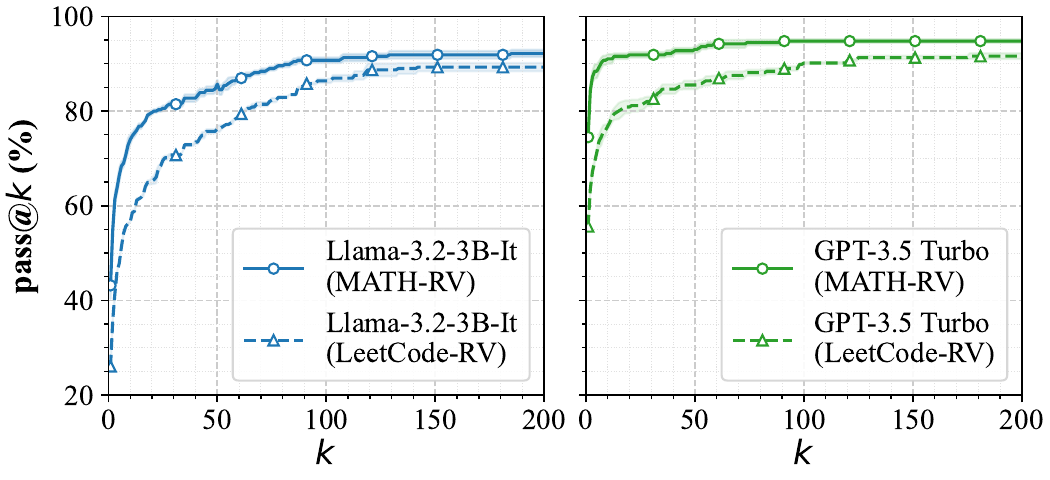}
    \caption{The accuracy evaluated using the $\text{pass@$k$}$ metric. Each line represents test-time scaling performance for a given LLM. The solid line denotes accuracy on \textsc{MATH-RV}, while the dotted line shows accuracy on \textsc{LeetCode-RV}.}
    \label{fig:test_time_scaling}
\end{figure}

Figure~\ref{fig:test_time_scaling} illustrates the $\text{pass@}k$ accuracy with multiple attempts.
Taking \textsc{LeetCode-RV} as an example, with a single attempt, Llama-3.2-3B-It achieves an accuracy of 26.67\%. However, with up to 10 attempts, its $\text{pass@}10$ accuracy increases substantially to 56.52\%.
Notably, the upper bounds of $\text{pass@}k$ performance in \textsc{LeetCode-RV} align closely with those in \textsc{MATH-RV}, reaching approximately 90\% at $\text{pass@}200$.
We identify the remaining 10\% of questions as likely representing high-difficulty questions (further discussion is provided in the extended version) that LLMs fail to solve due to fundamental limitations in their mathematical reasoning capabilities, possibly constrained in model size.
In other words, improving the performance on these high-difficulty questions remains beyond the reach of test-time scaling.
Apart from task difficulty, the degree of $\text{pass@}k$ improvement in the LeetCode distribution is noticeably greater than in MATH. 
We refer to this phenomenon as an \textit{\textbf{``elicitation of proficiency generalization in mathematical reasoning tasks''}}.
These findings provide indirect support for our earlier hypothesis: the inconsistency between \textsc{MATH-RV} and \textsc{LeetCode-RV} performance is more likely caused by \textbf{\textit{LLMs' imbalance in proficiency between encountered and ``unseen'' data distributions. The generalization of proficiency is not well-established across similar mathematical reasoning tasks but can be elicited by test-time scaling.}}

\section{Conclusions}
Motivated by significant limitations in existing mathematical reasoning benchmarks, such as simplistic design and potential data contamination, we introduce \textsc{RV-Bench}, a novel evaluation methodology that leverages RVQs to more accurately evaluate LLMs’ mathematical reasoning capabilities.
Our findings reveal substantial accuracy drops when LLMs encounter RVQs that are ``unseen'' during their training, highlighting the potential unreliability of existing benchmarks in truly capturing LLM performance.
While LLMs acquire partial mathematical proficiency during pre-training and fine-tuning, this proficiency is often tied to specific data distributions and exhibits limited generalization across broader mathematical tasks during evaluation.
However, we further demonstrate that test-time scaling can effectively elicit this generalization.
Overall, \textsc{RV-Bench} provides a more reliable and effective methodology for evaluating LLMs in mathematical reasoning, offering valuable insights for advancing LLM-based mathematical reasoning research and its application to real-world challenges.

\section{Acknowledgments}
The work described in this paper was fully supported by a grant from the Innovation and Technology Commission of the Hong Kong Special Administrative Region, China (Project No. ITS/263/24FP).

% \nocite{*}
\bibliography{aaai2026}

\newpage
\appendix
\section{Related Work}
\paragraph{Benchmarking LLMs.} The rapid development of LLMs has significantly advanced the evaluation of their capabilities~\cite{chang2024survey}.
Well-designed benchmarks such as MMLU~\cite{hendrycks2021mmlu}, GLUE~\cite{wang2018glue}, MMLU-Pro~\cite{wang2024mmlupro}, SuperGLUE~\cite{wang2019superglue}, CommonSenseQA~\cite{talmor2019commonsenseqa}, and ARC~\cite{clark2018think} have pioneered the evaluation of general tasks, including question answering, natural language understanding, and commonsense reasoning.
As LLMs achieve remarkable success across various downstream domains, there is a growing demand for task-specific benchmarks to assess their specialized capabilities~\cite{chang2024survey, chen2021evaluating, zuo2025medxpertqa}.
Consequently, an increasing number of domain-specific datasets have been introduced.
For instance, BoolQ~\cite{clark2019boolq} and SQuAD~\cite{rajpurkar2016squad} evaluate machine reading comprehension and language-based complex reasoning, while GSM8K~\cite{cobbe2021gsm8k}, MATH~\cite{hendrycks2021mathdataset}, MathQA~\cite{amini2019mathqa}, and MathBench~\cite{liu2024mathbench} focus on mathematical  problem-solving.
With the proliferation of LLMs, leaderboards such as the OpenLLM Leaderboard~\cite{beeching2024openllmleaderboardv1, fourrier2024openllmleaderboardv2} and OpenCompass~\cite{contributors2023opencompass} now offer comprehensive evaluation protocols across mainstream benchmarks.
To address more complex real-world reasoning tasks, recent benchmarks like InfiBench~\cite{li2024infibench}, R-Bench~\cite{guo2025rbench}, and OlympicArena~\cite{huang2024olympicarena} have been released for advanced evaluations.

Our proposed \textsc{RV-Bench} is a domain-specific benchmark designed to evaluate LLMs' mathematical reasoning capabilities.
Its random variables evaluation methodology provides an effective means of reflecting LLMs' genuine reasoning and understanding of mathematical problems.

\paragraph{LLMs in Mathematical Reasoning.}
Mathematical reasoning is a task that effectively reveals the capabilities of LLMs and has garnered significant attention within the research community, leading to remarkable advancements~\cite{ahn2024large, chen2025advancing}.
Techniques such as continual pre-training~\cite{lin2024not}, fine-tuning~\cite{yuan2023scaling}, and reinforcement learning~\cite{wang2024math} have been extensively employed to enhance LLMs' mathematical reasoning abilities.
These efforts have also driven the development of math-specific LLMs, including Qwen-Math~\cite{yang2024qwen25math}, DeepSeek-Math~\cite{shao2024deepseekmath}, and MetaMath~\cite{yu2024metamath}.
Consequently, advanced methodologies combined with ever-expanding training data have enabled LLMs to achieve near-perfect performance on complex mathematical reasoning benchmarks~\cite{fourrier2024openllmleaderboardv2}.
For example, OpenMath~\cite{toshniwal2024openmathinstruct} emphasized the importance of question diversity during fine-tuning, achieving 96\% accuracy on the grade-school arithmetic reasoning benchmark GSM8K~\cite{cobbe2021gsm8k}.
Similarly, OpenAI's o1~\cite{zhong2024evaluation} attained nearly 95\% accuracy on the high-school-level MATH benchmark~\cite{hendrycks2021mathdataset}.

Most recently, GSM-Symbolic~\cite{mirzadeh2024gsm} proposed limitations in LLMs' genuine arithmetic reasoning abilities by evaluating them on diverse questions generated from manually-designed symbolic templates of GSM8K~\cite{cobbe2021gsm8k} datasets.
The study showed that LLMs prefer to replicate memorized reasoning patterns encountered during training rather than genuine problem-solving in mathematical reasoning, posing a critical challenge to current evaluation methodologies.
To address this issue, we propose \textsc{RV-Bench}, which employs ``unseen'' random variable questions (RVQs) to effectively evaluate LLMs' mathematical reasoning.
By re-ranking and comparing LLM performance on the original MATH benchmark, we demonstrate that LLMs still face significant challenges in understanding and solving complex mathematical problems, thereby revealing their genuine reasoning capabilities.

\section{Construction of \textsc{RV-Bench}}
\label{sec:rvbencon}
In this section, we detail the construction of the question-generating functions described in the main content by thoroughly explaining the annotation process for each module. 
This section also serves as a guideline for \textsc{RV-Bench} annotation, aiming to facilitate future development of random variable-based baselines and dataset updating.

\subsection{Preliminaries}  
\paragraph{Annotators.} To ensure the quality of our benchmark, we publicly recruited 10 candidate annotators, all of whom are graduate students with strong backgrounds in mathematics and computer science (majoring in at least one domain and achieving an ``A'' grade in the other through postgraduate-level high school examination). 
Candidates were required to manually solve a selection of mathematical and coding problems (100 in total) sampled from the MATH test set and LeetCode for qualification purposes. 
Ultimately, six candidates with the highest examination grades were chosen to serve as the professional annotators for \textsc{RV-Bench}.

\paragraph{Raw Data Sampling.} We randomly and uniformly sample 250 benchmark problems from the MATH test set and 250 from the LeetCode-Math branch (500 in total), following the processing setup of PRM800K~\cite{lightman2024lets}. 

\subsection{Annotation Process}
For a specific problem, as exemplified in Figure 2, the annotator initially reviews the problem meticulously to ensure a thorough understanding of the original content and the text-based solution provided. 
If the problem is found to be unclear or insufficiently comprehended by the annotator, it is subjected to a post-calibration process for further discussion, details of which are provided in the calibration and post-filtering section. 
Problems that are well comprehended proceed to the annotation process as described below.

\begin{table}[!t]
\centering
\includegraphics[width=\linewidth]{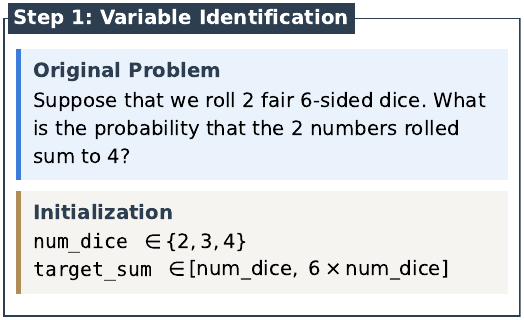}
\caption{\textbf{Step 1.} Annotators first review the problem and identify the variables, assigning a semantic-based name. Then, a range is set for each variable, which is required to maintain the difficulty level of the original problem.}
\label{tab:step1}
\end{table}

\paragraph{Step 1: Identify and Initialize the Variables.} 
For a problem meeting the criteria, the annotator begins by identifying the variables. 
Typically, key numbers, names, and equations are potential candidates for random variables. 
In practice, a problem may contain multiple variables, but to maintain a consistent difficulty level, only those variables that align with the original problem's conceptual intention are selected for randomization. 
As illustrated in Table~\ref{tab:step1}, the original problem involves calculating the probability for a specific sum with two dice. 
The number of dice and the target sum, which are integral to the problem's intent, are identified as variables, highlighted in green and yellow, respectively. 
The description ``6-sided dice'' which serves as a characteristic of the objective content, is not identified as a variable and is shown in gray. 
Once the variables are identified, the annotator locates their numerical and symbolic elements in the original problem and replaces them with slots.

Next, the annotator assigns a range to the identified variables, accompanied by semantically related variable names to define the \texttt{initialization} module. 
For variables that involve interdependent calculations, their ranges must be mutually constrained to ensure the question remains solvable (details are discussed in a later section). 
For example, in Tab.~\ref{tab:step1}, the range of the target sum depends on the number of dice. 
To initialize the question function, each variable is assigned a random value selected from its range.

\paragraph{Step 2: Construct the General Solution.}
For RVQs, the numerical outcomes will vary with different combinations of variable values. 
Therefore, a general solution must be constructed to solve the problem irrespective of these values.

In the case of problems from the MATH test set, the provided solutions include detailed text-based step-by-step problem-solving process. 
The annotator is required to convert these text-based solutions into code implementations. 
In certain instances, the annotator may need to revise the code implementation because some text-based solutions are coupled tightly to the specific problem and lack generalization. 
The process of constructing a general solution for LeetCode-Math problems differs slightly.
For each problem in LeetCode, some code-based solutions are available from the community, which have been validated through successful execution. 
The annotator must identify and comprehend a \texttt{Python} solution and transform it into the appropriate format, therefore defining the \texttt{solution} module based on the community-provided solutions.

\begin{table}[!t]
\centering
\includegraphics[width=\linewidth]{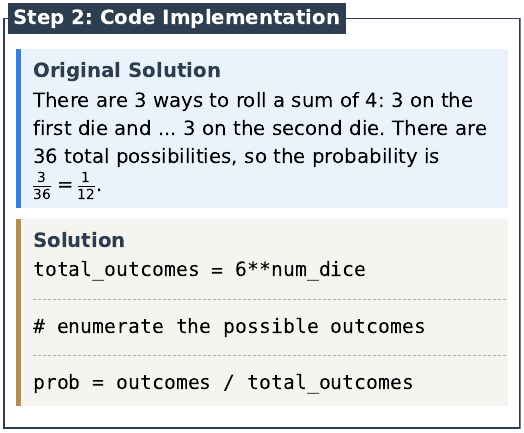}
\caption{\textbf{Step 2.} Annotators convert the original text-based step-by-step solution into code implementation. Due to the close coupling of some solutions with specific problems, annotators need to revise the solutions to ensure generalizable.}
\label{tab:step2}
\end{table}

Once the variable combination has been initialized in Step~\ref{tab:step1}, the \texttt{solution} module takes it as input and returns the corresponding answer. 
This module undergoes further validation for correctness and effectiveness through calibration among annotators, as detailed in the latter section.

\paragraph{Step 3: Generate the QA Pairs.} 
Following the completion of the previous steps, where the initialized variables and the general solution for various variable combinations were defined, annotators proceed to Step 3. 
In this step, annotators utilize the problem content with slots identified in Step 1 to define the \texttt{generation} module. 
This involves filling the slots with randomized variable values and formatting the output into question-answer (QA) pairs. Importantly, the original problem may include content that is extraneous to the problem (e.g., restrictions related to computational environments, useless formulas). 
Annotators are required to remove these irrelevant sections to ensure the question content focuses solely on the pertinent details.
All generated questions from different question functions are compiled into a comprehensive question set for \textsc{RV-Bench}.

\begin{table}[!t]
\centering
\includegraphics[width=\linewidth]{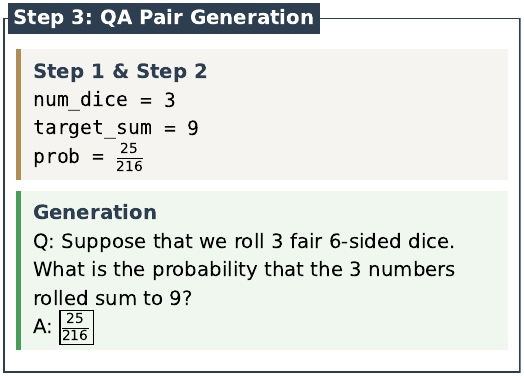}
\caption{\textbf{Step 3.} Based on the initialized variables from Step 1 and the corresponding answer from Step 2, the QA pair is generated as an instantiated question in \textsc{RV-Bench}.}
\label{tab:step3}
\end{table}

\subsection{Difficulty Control}
\label{sec:conditions}
To maintain the difficulty level of the RVQs consistent with the original problems, we incorporate difficulty control conditions when defining the random ranges for variables. 
We establish three conditions/criteria for setting the random range in Step~\ref{tab:step1}: 
1) The fluctuation range of variables should remain uniform across different questions; 
2) Variables that significantly affect the problem's complexity may be fixed as constants;  
3) The random range for simpler questions can be broader, whereas for more challenging questions, it should be narrower to prevent considerable variations in difficulty.

By controlling the difficulty level, we ensure that LLMs are fairly compared on RVQs relative to the original problem, minimizing performance differences that could arise from variations in difficulty. We conduct a detailed error analysis for LLMs in \textsc{RV-Bench} to demonstrate the effectiveness of difficulty control in the latter section.

\subsection{Calibration and Post-Filtering}
\label{sec:calibration_and_post-filtering}
Following the annotation process, we undertake a calibration and post-filtering step~\cite{li2024infibench} to enhance the consistency and objectivity of the question functions in RV-Bench. 
During Step 1, any problematic question that is not well-comprehended is promptly subjected to calibration and discussion. Confusing problems are collaboratively reevaluated and re-entered into the annotation process. 
If a problem cannot guarantee solvability or generalization for random variables, it is removed from the dataset.
After all question functions have been annotated, a cross-calibration process is conducted. 
Annotators review each other's annotations, verifying the correctness of the question functions and testing the runtime of the \texttt{solution} module across a broad spectrum of variable combinations. 
This runtime testing helps identify potential issues, such as exceeding maximum recursion depth, to ensure that each unique variable combination remains correctly solvable.
Additionally, question functions derived from LeetCode-Math that do not closely relate to mathematical reasoning or computation are filtered out.

\subsection{AI Tools in Annotation}
Regarding the annotation process, we employed LLMs such as GPT-4o and DeepSeek-V3 to generate initial code prototypes, which were subsequently refined by expert human annotators.
This integration of LLMs reduced the annotation time per item by approximately 15 minutes.
Nevertheless, we initially refrained from fully automating the process due to concerns about hallucinations and the reliability of LLMs in generating accurate code solutions for mathematical reasoning problems.
We will investigate this in future work.

\section{Experimental Setups}

\paragraph{Datasets.}
As a detailed extension of the main content, we provide additional information on the dataset used in the experimental section.
After calibration and post-filtering, \textsc{RV-Bench} comprises 230 question-generating functions.
For each function, we also sample the corresponding original problems from the MATH test set (\textbf{\textsc{MATH-Sp}}) and the LeetCode-Math branch (\textbf{\textsc{LeetCode-Sp}}).
For the MATH benchmark, the original problems are officially defined with both content and answer; thus, we directly adopt them for the \textsc{MATH-Sp} evaluation, using exact match accuracy as the metric.
For the newly introduced LeetCode-Math data source, we define the original problem for \textsc{LeetCode-Sp} as the version instantiated with the variable combinations and answers provided in the official example section below the problem description.
In total, 575 RVQs from 115 RVQ groups are generated by the MATH question-generating functions (\textbf{\textsc{MATH-RV}}), with an equal number generated by the LeetCode question-generating functions (\textbf{\textsc{LeetCode-RV}}), resulting in 1150 RVQs in \textsc{RV-Bench} for this study.
These two question sets, \textsc{MATH-RV} and \textsc{LeetCode-RV}, are used in all experiments and analysis throughout our study.
To ensure reproducibility, we use a fixed random seed of 42 for RVQ generation.
All generated RVQs and corresponding solutions will be publicly released.

\paragraph{Model Selection.} The selected models encompass a diverse range of LLMs, covering various model sizes and architectures to draw comprehensive conclusions across multiple aspects.
Given the current emphasis on open-source LLMs, we include widely-used representative models such as LLaMA~\cite{dubey2024llama3}, Qwen~\cite{bai2023qwen, yang2024qwen2}, Phi~\cite{abdin2024phi4}, Yi~\cite{young2024yi}, Gemma~\cite{team2024gemma2}, and DeepSeek~\cite{liu2024deepseek2}.
Beyond these dense-architecture models, we also incorporate mixture-of-experts (MoE) series LLMs, including Mixtral~\cite{jiang2024mixtral} and Qwen-Max.
In addition, we include math-specific models designed for domain expertise, such as Qwen-Math~\cite{yang2024qwen25math}, Mathstral~\cite{jiang2024mixtral}, and DeepSeek-Math~\cite{shao2024deepseekmath}.
For proprietary large-scale LLMs, we select well-known models like GPT-4o~\cite{achiam2023gpt4}, Claude-3~\cite{claud3report}, GLM-4-Plus~\cite{glm2024chatglm}, Gemini-2-Pro~\cite{team2024gemini}, and DeepSeek-V3~\cite{liu2024deepseek3}.
Finally, given the rapid advancements in large reasoning models (LRMs) from both academia and industry, to follow the advanced studies, we also include QwQ~\cite{qwen2024qwq, yang2024qwen2}, GLM-Zero, OpenAI's o1-preview/mini~\cite{qin2024o1}, o3-mini, and DeepSeek-R1~\cite{guo2025deepseekr1} in our \textsc{RV-Bench} evaluation.

\section{Further Analysis}
Beyond the main leaderboard results, we conduct a series of analyses to provide a deeper understanding of \textsc{RV-Bench} and to validate the soundness of our evaluation method.

\subsection{Macroscopic Analysis of \textsc{RV-Bench}}
\begin{figure}[!t]
    \centering
    \begin{subfigure}[b]{0.48\textwidth}
        \includegraphics[width=\textwidth]{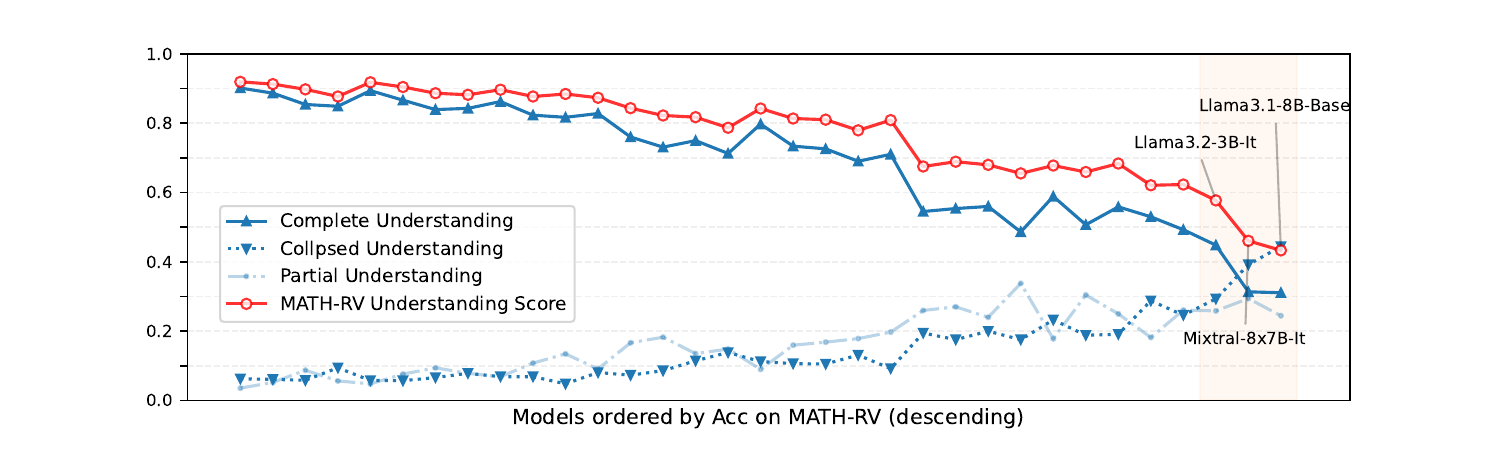}
        % \caption{Math}
        \label{fig:subfig_a}
    \end{subfigure}
    \\ \vspace{-0.15in}
    \begin{subfigure}[b]{0.48\textwidth}
        \includegraphics[width=\textwidth]{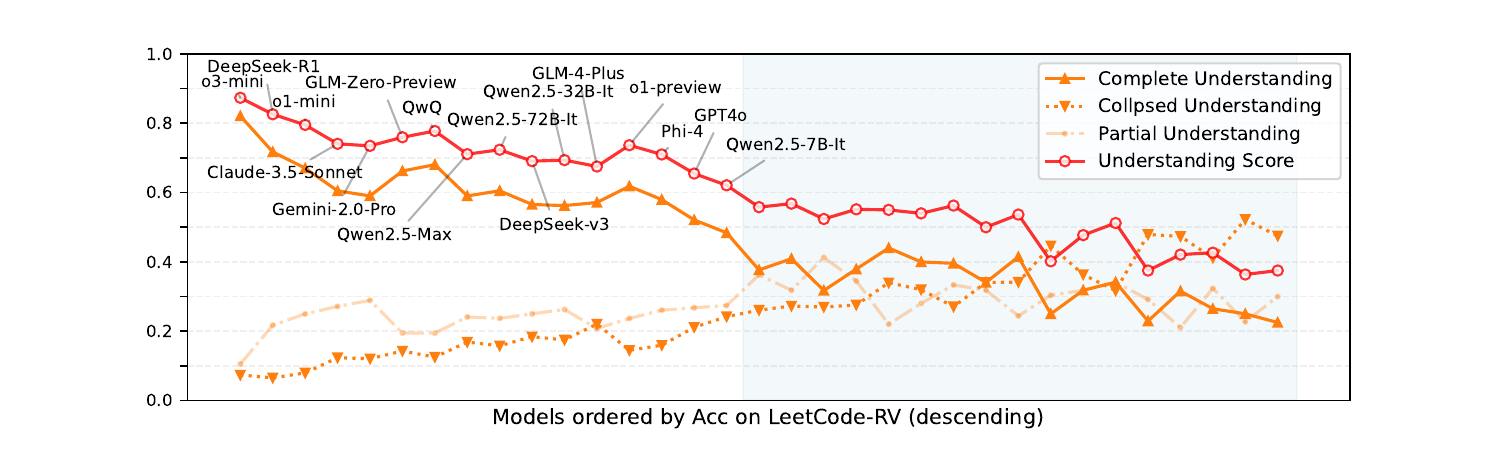}
        % \caption{Lc}
        \label{fig:subfig_b}
    \end{subfigure}
    \\ \vspace{-0.15in}
    \caption{
    The blue and orange lines show average pattern understanding degrees for correctly answered questions, while the red line (with circles) indicates the average pattern understanding score per RVQ group. Inconsistencies in the LLMs’ understanding are highlighted in the shaded regions.}
    \label{fig:com_coll}
\end{figure}

Figure~\ref{fig:com_coll} presents the average frequency of various understandings for each correctly answered original problem and the average pattern understanding score obtained by the corresponding RVQ group across different LLMs. As the direct visualization of the macroscopic analysis in main content.

\subsection{Error Analysis for Difficulty Control}
When evaluating LLMs using random variables, an intuitive concern arises: the observed performance gap between RVQs and original problems may not solely reflect the reasoning capabilities discussed in this study. Replacing fixed variables with random ones could inherently introduce additional numerical complexity and increase conceptual difficulty.
Given that LLMs are known to struggle with complex numerical computations, this concern may lead to significant accuracy drops, potentially undermining the completeness of our conclusions and the phenomena observed. 
To address this, the annotation process for question-generating functions enforces strict control over variable ranges, ensuring consistent numerical complexity, as outlined in the previous difficulty control section.
In this section, we provide experimental evidence to validate the effectiveness of this difficulty control and discuss its implications for our study.

We conduct an error analysis using five representative LLMs, selected based on their leaderboard rankings in \textsc{RV-Bench} and their prominence in the research community: DeepSeek-R1~\cite{guo2025deepseekr1}, GPT-4o~\cite{achiam2023gpt4}, Qwen2.5-7B~\cite{yang2024qwen2}, GPT-3.5 Turbo~\cite{tom2020language}, and Llama-3.2-3B~\cite{dubey2024llama3}.
For each model, we collect the incorrectly answered RVQs from \textsc{RV-Bench} and their corresponding original benchmark problems.
We then categorize the errors into three types for detailed analysis:

\noindent \textbf{1) Question Understanding Error (QUE)}: The model misunderstands the question requirement, leading to incorrect question understanding (e.g., the understanding of conditions, computational targets, answer format).

\noindent \textbf{2) Inference Logical Error (ILE)}: The inherent reasoning process of the given question is incorrect (e.g., misapplication of theorems, dimensional error, logic disalignment).

\noindent \textbf{3) Numerical Computational Error (NCE)}: The numerical computation is incorrect, resulting in an error in the final answer. This type of error is the primary focus of our error analysis by calculating the proportion difference.

\begin{table}[!t]
    \centering
    \footnotesize
    \begin{tabular}{lccc}
        \toprule
        \textbf{Model} & \textbf{QUE} & \textbf{ILE} & \textbf{NCE} \\
        \midrule
        \multicolumn{4}{>{\columncolor{sectionblue}}l}{\textbf{\textit{Original}}} \\
        DeepSeek-R1 & 30.43\% & 47.83\% & 21.74\% \\
        GPT-4o & 20.63\% & 52.38\% & 26.98\% \\
        Qwen2.5-7B-It & 29.41\% & 44.12\% & 26.47\% \\
        GPT-3.5 Turbo & 25.37\% & 49.68\% & 24.95\% \\
        Llama-3.2-3B-It & 31.62\% & 45.77\% & 22.61\% \\
        \midrule
        \addlinespace
        \multicolumn{4}{>{\columncolor{sectionblue}}l}{\textbf{\textit{Random Variables}}} \\
        DeepSeek-R1 & 32.51\% & 44.33\% & 23.15\% \textbf{(+1.41\%)} \\
        GPT-4o & 22.80\% & 52.02\% & 25.18\% \textbf{(-1.80\%)} \\
        Qwen2.5-7B-It & 27.08\% & 47.12\% & 25.80\% \textbf{(-0.67\%)} \\
        GPT-3.5 Turbo & 26.92\% & 47.85\% & 25.23\% \textbf{(+0.28\%)} \\
        Llama-3.2-3B-It & 31.04\% & 44.21\% & 24.75\% \textbf{(+2.14\%)} \\
        \bottomrule
    \end{tabular}
    \caption{Statistics of error types in proportion across models when answering Original benchmark problems and Random Variable Questions. The values in parentheses indicate the difference in the proportion between the original and random variable settings. The NCE is mostly consistent.}
    \label{tab:error_stats}
\end{table}

As shown in Table~\ref{tab:error_stats}, we tally the statistics of error types and calculate the difference in proportions between the original and random variable settings.
As observed, for the original benchmark problems, models already struggle with computational complexity, with NCE accounting for approximately 25\% of all errors.
When replacing fixed variables with random variables, the proportion of NCE remains at a similar level, fluctuating within 2\% of the original value.
This observation indicates that, despite altering numerical values, the design of \textsc{RV-Bench} does not significantly increase the numerical complexity of the questions.
Therefore, the conclusions and analyses presented in this paper are not confounded by potential increased numerical difficulty.

\begin{table}[!t]
    \centering
    \footnotesize
    \setlength{\tabcolsep}{4.6pt}
    \begin{tabular}{lccccc}
        \toprule
        \textbf{Model} & \textbf{$k{=}1$} & \textbf{$k{=}50$} & \textbf{$k{=}100$} & \textbf{$k{=}150$} & \textbf{$k{=}200$} \\
        \midrule
        \multicolumn{6}{>{\columncolor{sectionblue}}l}{\textbf{\textsc{MATH-RV}}} \\
        Llama-3.2-3B-It & 36.70 & 85.21 & 90.43 & 92.17 & 92.17 \\
        GPT-3.5 Turbo & 48.35 & 92.17 & 93.04 & 93.91 & 93.91 \\
        \midrule
        \multicolumn{6}{>{\columncolor{sectionblue}}l}{\textbf{\textsc{LeetCode-RV}}} \\
        Llama-3.2-3B-It & 23.83 & 76.52 & 83.48 & 89.57 & 90.43 \\
        GPT-3.5 Turbo & 31.48 & 79.13 & 84.35 & 90.43 & 92.17 \\
        \bottomrule
    \end{tabular}
    \caption{The change of $\text{pass@}k$ accuracy (\%) with the increase of the $k$ value for Llama-3.2-3B and GPT-3.5 Turbo on \textsc{MATH-RV} and \textsc{LeetCode-RV} datasetst.}
    \label{tab:passk}
\end{table}

Furthermore, these findings suggest that LLMs continue to face inherent challenges in solving certain types of mathematical problems, irrespective of perturbations introduced by variable combinations.
Additionally, when random variable combinations replace carefully designed values, LLMs may encounter non-integer intermediate computations, leading to further NCEs.
This phenomenon further highlights a fundamental weakness in LLMs' computational capabilities, particularly in handling precise numerical reasoning.

\subsection{Analysis of High-difficulty Questions}
As discussed in the Test-time Scaling Elicits Proficiency section of the main content, LLMs such as GPT-3.5 Turbo and Llama-3.2-3B can achieve accuracies approaching 90\% through test-time scaling.
We further asserted that ``the remaining 10\% of questions likely represent high-difficulty problems that LLMs fail to solve.''
Given the strength of this claim, this section provides a detailed analysis of these remaining questions. We first extend our experimental results with direct numerical representations, as shown in Table~\ref{tab:passk}.
Since higher accuracy corresponds to fewer remaining unsolved questions, and these remaining questions exhibit significant overlap, we select the remaining question set of Llama-3.2-3B for in-depth analysis.
As observed, as $k$ increases, the $\text{pass@}k$ accuracy for both \textsc{MATH-RV} and \textsc{LeetCode-RV} continues to improve, converging to approximately 90\% when $k=200$.

\begin{table}[!t]
    \centering
    \footnotesize
    \begin{tabular}{lccccc}
        \toprule
        \multicolumn{6}{c}{\textbf{Difficulty Levels}} \\
        \midrule
        \multicolumn{6}{>{\columncolor{sectionblue}}l}{\textbf{\textsc{MATH-RV}}} \\
        Level-1 & Level-2 & Level-3 & Level-4 & Level-5 & \\
        0.00\% & 0.00\% & 0.00\% & 11.11\% & 88.89\% & \\
        \midrule
        \multicolumn{6}{>{\columncolor{sectionblue}}l}{\textbf{\textsc{LeetCode-RV}}} \\
        \multicolumn{2}{l}{Easy} & \multicolumn{2}{l}{Medium} & Hard \\
        \multicolumn{2}{l}{0.00\%} & \multicolumn{2}{l}{18.18\%} & 81.82\%\\
        \bottomrule
    \end{tabular}
    \caption{Proportion of difficulty levels for the remaining unsolved questions by Llama-3.2-3B test-time scaling.}
    \label{tab:unsolved_difficulty}
\end{table}
To investigate the characteristics of these remaining unsolved questions, we conducted an in-depth error analysis and examined the distribution of question difficulty levels based on their original dataset annotations. 
As shown in Table~\ref{tab:unsolved_difficulty}, the difficulty levels are provided by the original datasets.
Our findings indicate that the unsolved questions under test-time scaling are predominantly those originally designed with high complexity, which exceed the current inherent reasoning capabilities of the models.

\section{Case Studies}
\begin{figure*}[!t]
\centering
\includegraphics[width=\textwidth]{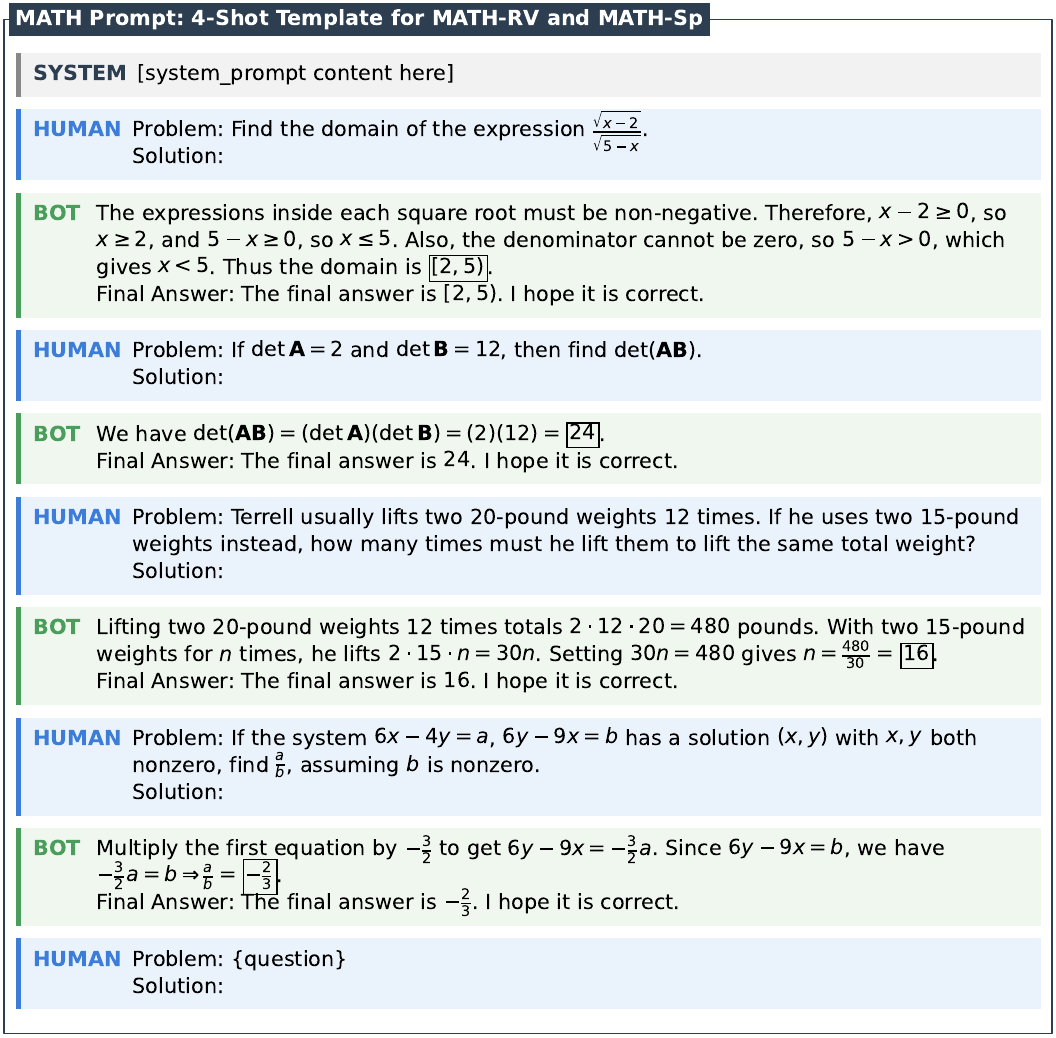}
\caption{4-Shot prompt template used in \textsc{MATH-RV} and \textsc{MATH-Sp} reasoning.}
\label{fig:math_prompt}
\end{figure*}

\begin{figure*}[!t]
\centering
\includegraphics[width=\textwidth]{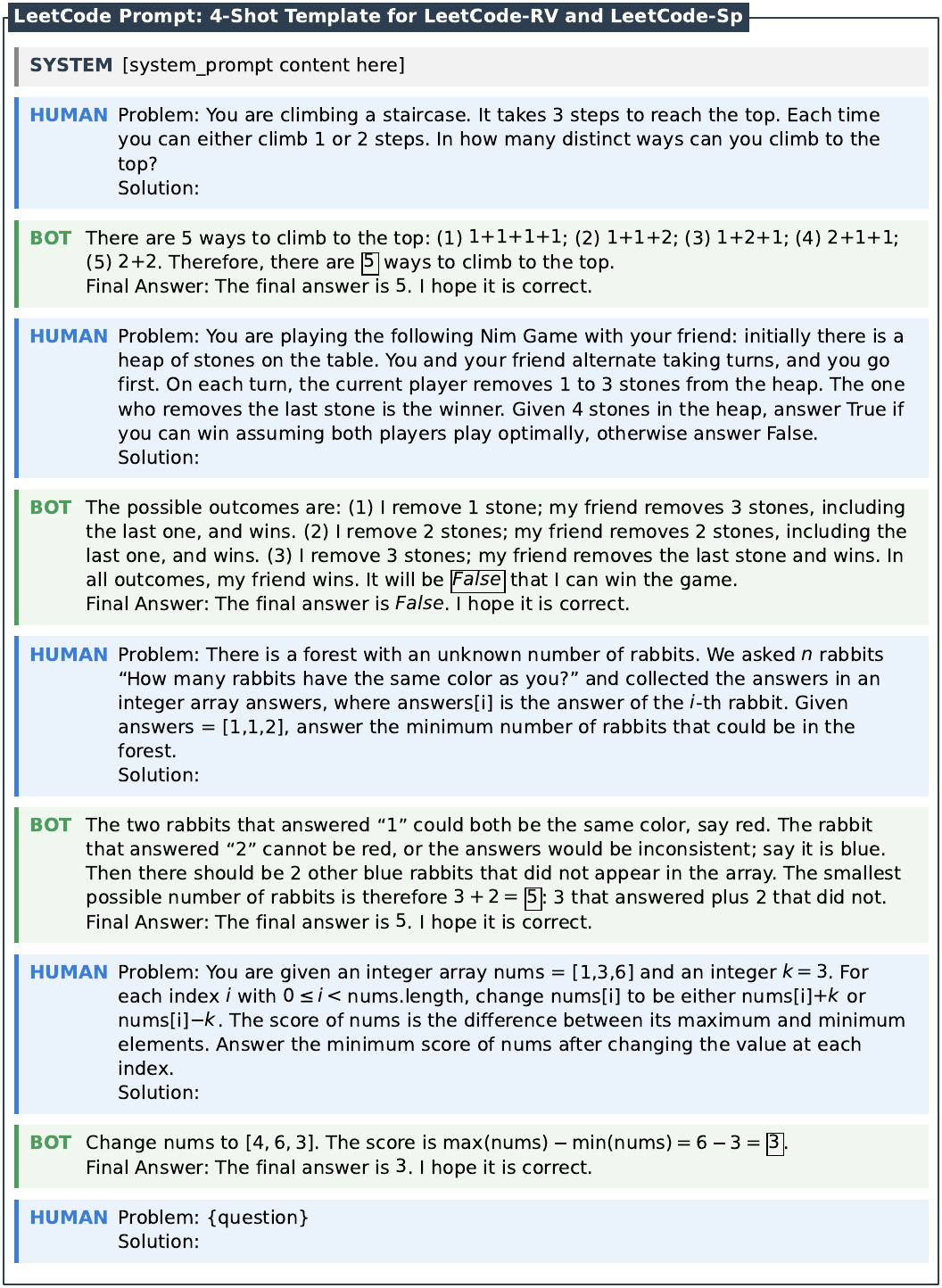}
\caption{4-Shot prompt template used in \textsc{LeetCode-RV} and \textsc{LeetCode-Sp} reasoning.}
\label{fig:leetcode_prompt}
\end{figure*}

In this section, we provide the prompt we used for \textsc{RV-Bench} evaluation.
For \textsc{MATH-RV} and \textsc{MATH-Sp} reasoning, we employ 4-shot prompting using problems from Minerva~\cite{lewkowycz2022solving}, which is the official setting provided by Llama series~\cite{dubey2024llama3}.
For \textsc{LeetCode-RV} and \textsc{LeetCode-Sp}, we randomly select four problems from LeetCode-Math out of \textsc{LeetCode-RV} and manually craft step-by-step solutions to serve as the few-shot examples.
The complete prompt templates are shown in Figure~\ref{fig:math_prompt} and Figure~\ref{fig:leetcode_prompt}.

\end{document}